\documentclass{article}

\usepackage{iclr2027_conference,times}

\usepackage{amsmath,amsfonts,bm}

\def\eqref#1{equation~\ref{#1}}

\def\1{\bm{1}}

\DeclareMathAlphabet{\mathsfit}{\encodingdefault}{\sfdefault}{m}{sl}
\SetMathAlphabet{\mathsfit}{bold}{\encodingdefault}{\sfdefault}{bx}{n}

\usepackage{amsmath}
\usepackage{amssymb}
\usepackage{booktabs}
\usepackage{array}
\usepackage{tabularx}
\usepackage{graphicx}
\usepackage{microtype}
\usepackage{xcolor}
\usepackage{url}
\usepackage{hyperref}
\hypersetup{hidelinks}

\title{AssayRouter: Historical Utility Priors for Frozen Molecular Predictor Routing}
\author{
\textbf{Dong Xu}$^{1,2}$, \textbf{Zhangfan Yang}$^{3}$, \textbf{Jiantao Wu}$^{1}$, \textbf{Shipeng Zhang}$^{1}$, \textbf{Zexuan Zhu}$^{1}$,\\
\textbf{Jiangjiang Li}$^{1}$, \textbf{Jun Zhang}$^{1}$, \textbf{Junkai Ji}$^{1,2}$\\[2mm]
\normalfont $^{1}$School of Artificial Intelligence, Shenzhen University\\
\normalfont $^{2}$EasternDawn\\
\normalfont $^{3}$School of Computer Science, University of Nottingham Ningbo
}
\iclrfinalcopy

\begin{document}
\maketitle
\fancyhead{}

\begin{abstract}
Laboratories often face a new molecular assay with 16--64 labels and a bank of predictors whose training data and parameters are unavailable. The practical question is which frozen outputs to include in a small local model. AssayRouter treats completed assays as pseudo-targets and labels each candidate by its post-fit utility: the reduction in held-out discovery loss when the candidate is added to the local target predictor. A shared regressor learns to predict this utility from candidate behavior on the support set, without source identity; on a new assay, one frozen ranking selects four sources and separate labels fit a convex combiner. We train only on completed ChEMBL-MT assays and evaluate 24 external regression assays across six frozen interface families. AssayRouter-C lowers strict four-call negative log-likelihood (NLL) by 0.0409 relative to Support-CV@4. Frozen candidate-label permutations confirm that candidate--utility correspondence carries the transferred information, and leave-one-interface-out training shows that the mapping generalizes to unseen predictor families. Completed assays therefore provide transferable supervision for scarce-label routing through frozen prediction interfaces.
\end{abstract}

\section{Introduction}

Drug discovery programs accumulate predictors assay by assay. When a new continuous endpoint has few measurements, models trained for other assays can already score the same molecules. Their examples, architectures, and parameters may be inaccessible. Reuse then occurs through a frozen contract: each source exposes predictions and metadata, while its training data and parameters remain unavailable. The practical decision is which output channels should enter a small model fitted to the target.

Recent few-shot methods adapt or condition a predictor on a small labeled set: ActFound uses pairwise meta-learning~\citep{feng2024actfound}, while FS-CAP~\citep{eckmann2024fscap} and UniMatch~\citep{li2025unimatch} encode labeled support. Other systems select or weight source tasks, as in GATE~\citep{lee2024gate}, or source data, as in AssayMatch~\citep{fan2026assaymatch}. These systems select data, tasks, or trainable models without labeling the downstream contribution of a deployed frozen prediction channel.

A source can be accurate in isolation yet redundant after fitting the local target model; a weaker source can correct a consequential residual. The useful supervision is therefore incremental value inside the fitted combination, which completed assays make observable. \textsc{AssayRouter} measures how much each frozen source reduces held-out loss after the same local fit used at deployment. A regressor maps candidate behavior on the support set to this post-fit utility, conditioned on the interface family. Frozen on completed assays, it gives an unseen assay one score per source, one Top-4 set, and separately fitted convex weights.

The evaluation crosses 24 external assays in three collections with six interface families. Matched interventions test whether candidate-specific utility transfers, whether it survives removal of an interface family, and whether frozen ranking can replace subset search on the target. These experiments support three contributions:
\begin{itemize}
    \item We define post-fit utility, the incremental loss reduction from adding one frozen source to a local target predictor, as transferable supervision for routing.
    \item AssayRouter amortizes subset search into one frozen score per source, replacing more than 20 million inner nonnegative least-squares (NNLS) fits with zero counterfactual fits on the target.
    \item Controlled interventions show that candidate-specific utility transfers across external assays and to predictor families absent from router training, improving strict four-call routing over Support-CV@4.
\end{itemize}

\section{Problem Setting}
\label{sec:setting}

Let an unseen target assay $t$ provide labeled support $D_t^{\mathrm{sup}}=\{(x_i,y_i)\}_{i=1}^{n}$ and an untouched confirmation set $D_t^{\mathrm{test}}$. A bank $\mathcal S_t$ contains independently trained source predictors $f_s$. Each interface exposes a prediction $z_s(x)$, optional uncertainty $u_s(x)$, and declared metadata $m_s$, while its training data and parameters remain inaccessible. We divide each support episode into balanced, disjoint roles: routing observes candidate behavior and metadata on $R_t$, while fitting observes the selected source outputs on $C_t$. The two operations receive separate contracts:
\begin{equation}
\mathcal I_t^{\mathrm{route}}=
\left(R_t,\{z_s(R_t),u_s(R_t),m_s\}_{s\in\mathcal S_t},\iota_t\right),\quad
\mathcal I_t^{\mathrm{fit}}(A_t)=
\left(C_t,\{z_s(C_t)\}_{s\in A_t}\right),\quad R_t\cap C_t=\varnothing.
\label{eq:information-contract}
\end{equation}
where $\iota_t$ denotes the deployed interface family. In the evaluated contract, $m_s$ includes source training size, while $z_s$ and $u_s$ are the mean and empirical dispersion of each source's frozen members. The confirmation set is absent from both contracts and is opened only for final scoring. The target is also deleted from source construction, router training, model selection, and calibration; historical episodes apply the same deletion.

The task contains two decisions. A selector chooses $A_t\subset\mathcal S_t$ with $|A_t|=K=4$; a normalized NNLS combiner then fits weights on $C_t$ for the local target column and four selected source outputs. This fan-in defines the call budget at confirmation and keeps target adaptation identifiable when labels are scarce. For any legal set $A$, let $\widehat h_{t,A}$ denote that fitted combiner and write $J_t(A)=\mathcal L_{D_t^{\mathrm{test}}}(\widehat h_{t,A})$. The latent optimal action at confirmation and selection regret are
\begin{equation}
A_t^*=\arg\min_{A\subseteq\mathcal S_t,\,|A|=K}J_t(A),\qquad
\mathcal R_t(\pi)=J_t\!\left(\pi(\mathcal I_t^{\mathrm{route}})\right)-J_t(A_t^*).
\label{eq:selection-regret}
\end{equation}
The oracle action is unavailable at deployment because confirmation labels remain sealed. Since $J_t(A_t^*)$ is shared across selectors, paired differences in confirmation loss equal paired differences in latent regret. AssayRouter learns a policy $\pi:\mathcal I_t^{\mathrm{route}}\mapsto A_t$ from completed assays, then fits $\widehat h_{t,A_t}$ through the separate fitting contract. The study asks whether post-fit utility learned from completed assays reduces this latent regret on a new label function under a frozen interface family.

\section{AssayRouter}
\label{sec:method}

\subsection{Learning a post-fit utility prior}

\begin{figure}[ht]
\centering
\includegraphics[width=\linewidth]{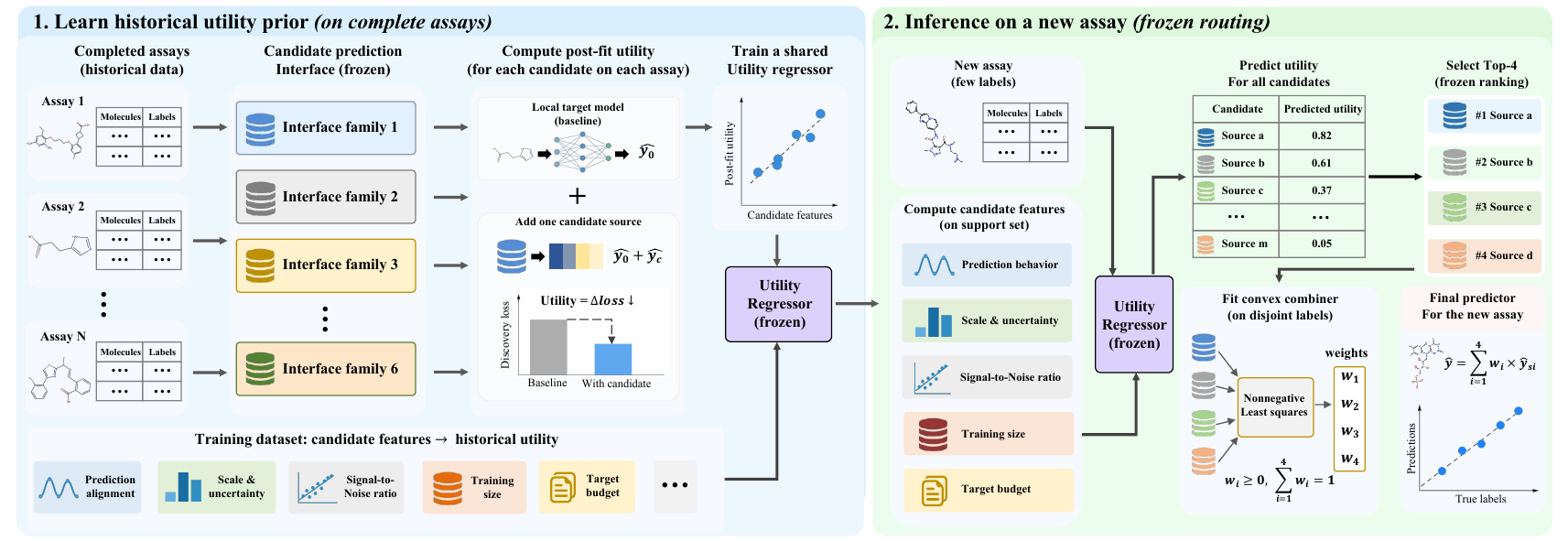}
\vspace{-8mm}
\caption{\textbf{Completed assays supervise frozen predictor routing.}
(1) A historical pseudo-target measures the reduction in discovery loss from adding one
candidate. (2) A shared AssayRouter prior maps candidate behavior to utility without
source identity. (3) On a new assay, the frozen prior scores the bank once, selects four
sources, and fits simplex weights for the target before confirmation.}
\label{fig:overview}
\vspace{-2mm}
\end{figure}

Figure~\ref{fig:overview} connects the three stages: measuring post-fit utility on completed assays, learning one source prior without identity features, and freezing a four-source action before fitting the target.

We measure post-fit utility by mirroring deployment on each completed assay: fit a convex combination of the local predictor and candidate sources, then evaluate it on held-out discovery molecules. Let $e$ index such a historical episode and $A\subseteq\mathcal S_e$ a candidate subset. A four-fold out-of-fold Ridge model supplies the local target column $\widehat y_{0,e}^{\mathrm{CF}}$. With $\widetilde X_{e,A}(x)=[\widehat y_{0,e}^{\mathrm{CF}}(x),(z_s(x))_{s\in A}]$, fit $\widetilde\beta_{e,A}$ and normalize it as $\widehat\beta_{e,A}=\widetilde\beta_{e,A}/(\mathbf1^\top\widetilde\beta_{e,A})$:
\begin{equation}
\widetilde\beta_{e,A}=\operatorname{argmin}_{\beta\geq0}
\|\widetilde y_{C_e}-\widetilde X_{e,A}(C_e)\beta\|_2^2,
\qquad
\widehat h_{e,A}(x)=\mu_{C_e}+\sigma_{C_e}
\widetilde X_{e,A}(x)\widehat\beta_{e,A}.
\label{eq:simplex-combiner}
\end{equation}
Tildes denote the declared $C_e$-local robust scaling; $\widehat h$ is returned to the raw response scale before evaluation. If every NNLS coefficient vanishes, we use the uniform simplex vector. For discovery set $Q_e^{\mathrm{disc}}$, disjoint from $R_e$ and $C_e$, write $\mathcal L_{e,A}=\mathcal L_{Q_e^{\mathrm{disc}}}^{(\sigma_{C_e})} (\widehat h_{e,A})$ and define
\begin{equation}
\mathcal L_Q^{(\sigma)}(h)=|Q|^{-1}
\sum\nolimits_{(x,y)\in Q}
\ell_\nu\!\bigl((y-h(x))/\sigma\bigr),
\quad
\Delta_e(s\mid A)=\mathcal L_{e,A}-\mathcal L_{e,A\cup\{s\}},
\label{eq:set-utility}
\end{equation}
where the Student-$t$ loss for point predictions with $\nu=3$ is
\begin{equation}
\ell_\nu(r)=\tfrac12\log(\nu\pi)+\log\Gamma(\nu/2)
-\log\Gamma((\nu+1)/2)+\tfrac{\nu+1}{2}\log(1+r^2/\nu).
\label{eq:student-loss}
\end{equation}
AssayRouter measures the post-fit value $U_e(s)=\Delta_e(s\mid\varnothing)$. Its raw level includes a block-wide offset induced by historical target difficulty and support quality, so we center the candidates within each block:
\begin{equation}
U_e^{\mathrm C}(s)=U_e(s)-|\mathcal S_e|^{-1}\sum\nolimits_{r\in\mathcal S_e}U_e(r)
=|\mathcal S_e|^{-1}\sum\nolimits_{r\in\mathcal S_e}\mathcal L_{e,\{r\}}-\mathcal L_{e,\{s\}}.
\label{eq:centered-utility}
\end{equation}
Centering removes the shared offset while preserving every candidate ordering (Eq.~\ref{eq:centered-ranking-invariance}). AssayRouter-$\Delta$ retains the uncentered loss reduction as a matched calibration ablation.

A single histogram gradient-boosting (HGB) regressor predicts centered utility from candidate behavior, conditioned on the interface family. Source and assay identity never enter the regressor, and its representation reads neither $C_e$ nor $Q_e^{\mathrm{disc}}$:
\begin{equation}
\widehat\theta_\lambda=\operatorname{argmin}_{\theta}
\sum\nolimits_e\sum\nolimits_{s\in\mathcal S_e}
w_{e,s}\left[q_\theta\!\left(\psi_e(s;R_e),\iota_e\right)-U_e^{\mathrm C}(s)\right]^2.
\label{eq:router-objective}
\end{equation}
We choose $\widehat\lambda\in\Lambda$ by cross-validation mean absolute error (MAE) averaged over targets; grouped folds keep every interface of a historical assay on the same side. Appendix~\ref{app:frozen-fitting} gives the configurations and training rule, while Appendix~\ref{app:feature-inventory} defines the repeated-state weights $w_{e,s}=13$.

The 11 coordinates separate complementarity from reliability. Signed and absolute residual alignment and residual correlation measure how a source tracks what the local model misses; source scale, spread, uncertainty, signal-to-noise ratio (SNR), training size, and routing-support size describe whether that signal is reliable. Let $r_{e,i}=\widetilde y_{e,i}-\widehat y_{0,e}^{\mathrm{CF}}(x_i)$ and $v_{s,i}=\widetilde z_s(x_i)$ on $R_e$. The primary representation is
\begin{equation}
\begin{aligned}
\psi_e(s;R_e)=\big(&
\overline{v_sr_e},\ |\overline{v_sr_e}|,\ \operatorname{corr}(v_s,r_e),
\ \overline{v_s},\ \operatorname{sd}(v_s),\ \overline{|v_s|},\\
&\overline{u_s},\ \operatorname{sd}(u_s),
\ \operatorname{sd}(v_s)/\max(\overline{u_s},10^{-3}),
\ \log(1+n_s),\ \log_2|R_e|\big).
\end{aligned}
\label{eq:routing-features}
\end{equation}
All coordinates are available under the routing contract; the interface code is appended separately.

Equations~\ref{eq:simplex-combiner}--\ref{eq:router-objective} define the name \emph{AssayRouter}: completed assays convert post-fit candidate contribution into a shared routing prior, and the prior turns a new support set into one frozen source allocation. The centering step changes calibration without changing the supervised action,
\begin{equation}
\operatorname{TopK}_{s\in\mathcal S_e}U_e^{\mathrm C}(s)
=\operatorname{TopK}_{s\in\mathcal S_e}U_e(s),
\label{eq:centered-ranking-invariance}
\end{equation}
while making utility levels comparable across historical blocks.

\noindent\textbf{End-to-end procedure.} Learning and deployment follow four operations with separate data roles.
{\setlength{\leftmargini}{2.5em}
\begin{enumerate}
\item For each historical episode, fit the local target Ridge column on $R_e$ by
cross-fitting and compute $\psi_e(s;R_e)$ for every frozen candidate.
\item On $C_e$, fit the target-only model and its candidate-augmented counterpart from
Equation~\ref{eq:simplex-combiner}; on disjoint $Q_e^{\mathrm{disc}}$, convert their
loss difference into $U_e^{\mathrm C}(s)$.
\item Pool historical episodes and fit $\widehat q$ with model selection grouped by assay.
Freeze its feature schema, interface encoding, capacity, and parameters.
\item For an unseen target, recompute candidate behavior on $R_t$, select $A_t$ once,
and fit only $\widehat\beta_{t,A_t}$ on $C_t$. Confirmation labels enter the final
metric after both the set and weights have been fixed.
\end{enumerate}
}
Thus $R$ determines the routing representation, $C$ identifies the local convex combiner, and historical $Q_{\mathrm{disc}}$ supplies the utility label. No deployment operation reconstructs the discovery query or revisits the selected set.

Equation~\ref{eq:set-utility} operationalizes this incremental value under the same convex refit used at deployment. The population target is the component of post-fit value predictable from a new routing support:
\begin{equation}
q^*(\psi,\iota)=\mathbb E\!\left[U_e^{\mathrm C}(s)\mid
\psi_e(s;R_e)=\psi,\ \iota_e=\iota\right].
\label{eq:population-router}
\end{equation}
Leave-one-interface-out training additionally removes $\iota$ for the held-out family and tests the behavioral mapping on an unseen representation.

\noindent\textbf{Nested refinements.} Let $G_{e,s}=\mathcal G_e(s)$ be the 13 frozen partial sets of sizes zero through three that exclude $s$, and abbreviate $Q_e(s,A)=q_{\mathrm S}(\psi_e(s;R_e),g(A),\iota_e)$ and $Q_{e,j-1}(s)=Q_e(s,A_{j-1})$, with $\mathcal C_j=\mathcal S_e\setminus A_{j-1}$. The marginal-over-states label and state-conditioned action are
\begin{equation}
U_e^{\mathrm M}(s)=\operatorname{Avg}_{A\in G_{e,s}}\Delta_e(s\mid A),\quad
s_j=\operatorname{argmax}_{s\in\mathcal C_j}Q_{e,j-1}(s),\quad
A_j=A_{j-1}\cup\{s_j\}.
\label{eq:nested-refinements}
\end{equation}
Here $\operatorname{Avg}$ is the arithmetic mean and $g(A)$ contains 14 set-state coordinates. Both refinements retain the historical rows and external replay; they change the supervision or state representation only.

\noindent\textbf{Mechanism interventions.} Within each historical target--interface--budget--episode block, a bijection permutes utility among the 19 candidates while preserving feature rows, the label multiset, capacity search, grouped folds, and external assignments. Six leave-one-interface-out routers remove one predictor family and its calibration from historical training.

\subsection{Routing once with the frozen utility prior}

\noindent\textbf{Frozen action and target-specific fit.} At deployment, the frozen router selects four candidates from one score per source:
\begin{equation}
A_t=\operatorname{TopK}_{s\in\mathcal S_t}
\widehat q\!\left(\psi_t(s;R_t),\iota_t\right),
\qquad |A_t|=K=4.
\label{eq:one-shot-routing}
\end{equation}
Routing reads only $\mathcal I_t^{\mathrm{route}}$; after $A_t$ is frozen, Equation~\ref{eq:simplex-combiner} fits its weights on the separate $C_t$ contract. Writing $z_0(x)=\widehat y_{0,t}(x)$, the confirmation predictor is
\begin{equation}
\widehat y_t(x)=\sum\nolimits_{j\in\{0\}\cup A_t}\widehat\beta_{t,j}z_j(x),
\qquad \widehat\beta_{t,j}\geq0,\quad
\sum\nolimits_{j\in\{0\}\cup A_t}\widehat\beta_{t,j}=1.
\label{eq:deployed-predictor}
\end{equation}
Only membership and the fitted simplex weights enter prediction. For $a>0$ and any constant $c$, the action is affine-invariant:
\begin{equation}
\operatorname{TopK}(a\widehat q+c)=\operatorname{TopK}(\widehat q)=A_t.
\label{eq:affine-action-invariance}
\end{equation}
Score calibration is therefore irrelevant until it changes the Top-4 boundary. Each confirmation molecule calls the four selected sources. Across all 24 external targets, the learned prior, feature schema, and interface encoding remain fixed; each target recomputes its local predictor and routing features, then determines $A_t$ and $\widehat\beta_{t,A_t}$.

\noindent\textbf{Amortized selection cost.} With $M$ sources, AssayRouter performs $M$ frozen score evaluations, one top-$K$ operation, and no counterfactual fit on the target. A Support-CV screen of width $L$ and $F$ inner folds instead uses $B=\min(M,L)$ screened sources, where $\operatorname{Comb}(B,K)$ counts their $K$-source subsets:
\begin{equation}
N_{\mathrm{score}}^{\mathrm{AR}}=M,\qquad
N_{\mathrm{fit}}^{\mathrm{AR}}=0,\qquad
N_{\mathrm{fit}}^{\mathrm{CV}}=
F\!\left(M\mathbf1\{M>L\}+\operatorname{Comb}(B,K)\right).
\label{eq:selection-workloads}
\end{equation}
The indicator term is paid only when $M>L$: each fold first scores all $M$ candidates to retain $B$. The combination term then fits every $K$-source subset of those $B$ survivors, and the factor $F$ repeats this selection workload across inner folds. Both methods subsequently fit the same final NNLS combiner with five columns. Across the full evaluation, Equation~\ref{eq:selection-workloads} yields 20,445,696 Support-CV inner fits and zero for AssayRouter; Appendix~\ref{app:amortized-cost} gives the ledger by collection and the crossover. This count excludes inference by frozen sources and the common final NNLS fit, isolating the work on the target used to decide membership. Historical utility construction is paid once and reused across external assays, so repeated deployment increases Support-CV search cost while leaving router fitting fixed.

\noindent\textbf{Two evaluation estimands.} Let $\Pi_t$ contain the 16 constituents formed by eight balanced $R/C$ partitions and both role directions, with $\widehat h_{t,\pi}$ the frozen four-source predictor for constituent $\pi$. On confirmation set $D_t^{\mathrm{test}}$, define
\begin{equation}
\mathcal L_{D_t^{\mathrm{test}}}(h)=
|D_t^{\mathrm{test}}|^{-1}
\sum\nolimits_{(x,y)\in D_t^{\mathrm{test}}}
\ell_3\!\bigl((y-h(x))/\sigma_{R_t\cup C_t}\bigr),
\label{eq:confirmation-loss}
\end{equation}
where $\sigma_{R_t\cup C_t}$ is the robust scale of the complete labeled support. With $\widehat y_t^{\mathrm{CF}}(x)=16^{-1}\sum_{\pi\in\Pi_t}\widehat h_{t,\pi}(x)$, the prediction-averaged and strict estimands are
\begin{equation}
\widehat{\mathcal L}_t^{\mathrm{CF}}=
\mathcal L_{D_t^{\mathrm{test}}}(\widehat y_t^{\mathrm{CF}}),\qquad
\widehat{\mathcal L}_t^{\mathrm{strict}}=16^{-1}
\sum\nolimits_{\pi\in\Pi_t}
\mathcal L_{D_t^{\mathrm{test}}}(\widehat h_{t,\pi}).
\label{eq:deployment-estimands}
\end{equation}
Cross-fit scores one variance-reduced prediction averaged over 16 constituents and can touch 5--21 unique sources per episode. Strict scoring averages 16 scalar losses after each constituent has made exactly four calls; it is therefore the primary deployment estimand. All sparse methods share the assignments, bank, fan-in, and final convex fit. Support-CV@4 searches using support labels before confirmation; Frozen-DES@4 (dynamic ensemble selection) and MINE-WS@4~\citep{moura2021mine} receive query-local access, and all-source convex relaxes fan-in. Random and residual diagnostics remain in the appendix.

\noindent\textbf{Action stability.} Let $\widehat q_{(K)}$ and $\widehat q_{(K+1)}$ bracket the selected set. For any refined scorer $q'$,
\begin{equation}
\gamma_t=\widehat q_{(K)}-\widehat q_{(K+1)},\qquad
\max_{s\in\mathcal S_t}|q'(s)-\widehat q(s)|<\gamma_t/2
\quad\Longrightarrow\quad
\operatorname{TopK}(q')=A_t.
\label{eq:selection-margin}
\end{equation}
A refined score can change the selected set only by crossing this rank margin. Whether that change alters prediction depends on the refitted combination function, including the displaced source and all recomputed weights.

\section{Results}
\label{sec:results}

We fit the router and select its HGB capacity on 20 ChEMBL-MT pseudo-targets, then evaluate a fixed four-source contract on 24 external targets across six interfaces, three label budgets, and 12 support replays. ExpansionRx~\citep{openadmet2026expansionrx} provides temporal holdouts, Biogen~\citep{fang2023prospective} sparse continuous endpoints, and TDC~\citep{huang2021tdc} scaffold-split regression tasks. Earlier sourcewise studies selected the contract (Appendix~\ref{app:retrospective-development}); the primary sparse comparisons share its deployment budget. At 16 labels, each directional fitting split contains eight observations for the local predictor and four selected channels, making reliable subset search especially difficult.

For target $t$, interface $\iota$, budget $b$, and replay set $\mathcal E_{t\iota b}$, every comparison is first paired at the episode level and then aggregated as
\begin{equation}
d_{t\iota b}=\frac1{|\mathcal E_{t\iota b}|}
\sum_{e\in\mathcal E_{t\iota b}}
\left(\widehat{\mathcal L}^{\mathrm{comp}}_{e}-
\widehat{\mathcal L}^{\mathrm{AR}}_{e}\right).
\label{eq:paired-cell-effect}
\end{equation}
Aggregate intervals independently resample target and interface clusters, so support partitions are not treated as independent scientific replicates; Appendix~\ref{app:set-statistics} gives the full unit counts.

The comparisons isolate bank reuse, utility supervision, ranking signal, information timing, and action fan-in while retaining the same frozen bank, $R/C$ roles, Top-4 action, and final NNLS fit whenever the tested axis permits. Target-only tests source reuse; source-quality, support statistics, residual compatibility, and permutations test the ranking signal; Support-CV@4 tests search on the target. Query-local and dense controls deliberately relax access or fan-in. Appendix~\ref{app:reuse-paradigms} positions every comparator, and Table~\ref{tab:information-budget-main} records its permissions. Support-standardized Student-$t$ NLL is primary; MAE, root mean squared error (RMSE), and Spearman correlation provide complementary error and ranking views.

\subsection{AssayRouter transfers post-fit utility}

\begin{figure}[ht]
\centering
\includegraphics[width=0.98\textwidth]{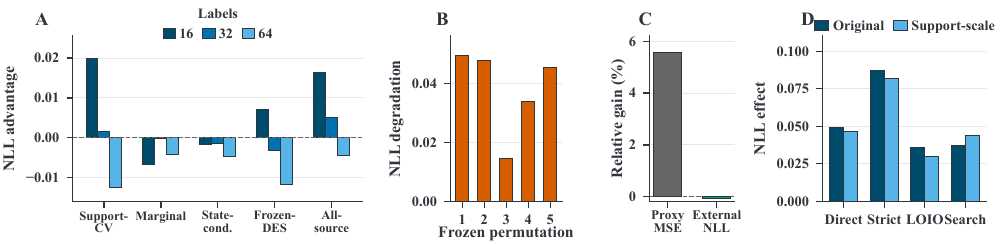}
\vspace{-4mm}
\caption{\textbf{Post-fit utility transfers across assays.}
(A) Comparator-minus-AssayRouter NLL at each target-label budget; positive values
favor the utility prior.
(B) Five independently frozen candidate-label permutations each degrade external NLL.
(C) Direction-normalized proxy MSE and external NLL changes; positive denotes improvement and exposes their divergence.
(D) Original and support-scale versions of the direct, strict, unseen-interface, and matched-search comparisons.}
\vspace{-2mm}
\label{fig:set-conditional-results}
\end{figure}

Candidate identity must remain paired with its historical post-fit value for routing to transfer. Five frozen candidate-label permutations each degrade aggregate external NLL under prediction averaging (Fig.~\ref{fig:set-conditional-results}B). The effect is positive at every budget and across all six interfaces in the uncentered replication (Table~\ref{tab:standalone-permutation-intervals}). The intervention preserves feature rows, the utility multiset, model capacity, grouped folds, and external assignments. Under strict scoring, the same permutation increases AssayRouter-C's NLL by 0.1352 (95\% CI $[0.0852,0.1915]$), favoring the genuine mapping in 90.7\% of paired cells. The advantage is largest at 16 labels, when evidence on the target is scarcest (Fig.~\ref{fig:set-conditional-results}A). It persists with uncentered labels (Table~\ref{tab:strict-permutation-nll}) and support-scale normalization (Table~\ref{tab:qdisc-scale-sensitivity}).

\vspace{-2mm}

\begin{figure}[ht]
\centering
\begin{minipage}[c]{0.64\textwidth}
\includegraphics[width=\linewidth]{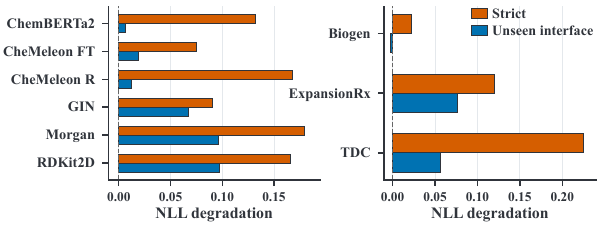}
\end{minipage}\hfill
\begin{minipage}[c]{0.32\textwidth}
\caption{\textbf{Candidate--utility transfer spans interfaces and collections.}
Mean NLL degradation after label permutation by deployed interface (left) and
external collection (right); both interventions use AssayRouter-C, and positive
values favor the genuine post-fit labels.}
\label{fig:transfer-scope}
\end{minipage}
\vspace{-2mm}
\end{figure}

The learned correspondence extends beyond the identities available during historical training. The strict intervention retains every interface family; leave-one-interface-out training removes the evaluated family. With AssayRouter-C, permuting the transferred correspondence increases NLL by 0.0497 with a clustered 95\% interval of $[0.0135,0.0963]$ and positive point effects in all six held-out families (Fig.~\ref{fig:transfer-scope}). The same unseen-interface routers improve over Support-CV by 0.1116 NLL with interval $[0.0834,0.1428]$ and win on all 24 targets (Table~\ref{tab:loio-permutation}).

A crossed audit excluding both historical target and candidate-source identities retains utility accuracy and Top-4 selection (Fig.~\ref{fig:acceptance-controls}, right; Appendix~\ref{app:source-identity}). Exact-parent removal on TDC preserves both effects in the uncentered replication (Table~\ref{tab:compound-nonoverlap}). Table~\ref{tab:evaluation-matrix} records the evaluation boundary.

\vspace{-2mm}

\subsection{Historical utility outperforms selection on the target under the same action}

The strict estimand asks the deployment question directly: after one frozen Top-4 choice, how well does each four-call constituent predict? AssayRouter-C gives the best overall NLL, MAE, and RMSE among the access-matched methods, with its advantage largest at 16 labels (Table~\ref{tab:strict-four-call-main}). The $\Delta$, Z, and marginal-over-states variants remain close, making post-fit candidate value the shared supervision principle across these label constructions.

\begin{table}[!t]
\centering
\caption{\textbf{Strict four-call deployment.} C, $\Delta$, and Z denote centered,
uncentered, and standardized standalone utility; Marginal denotes marginal-over-states utility. Every row uses one-shot
constituent scoring; bold and underline mark the two best point estimates.}
\label{tab:strict-four-call-main}
{\small
\setlength{\tabcolsep}{4pt}
\begin{tabular}{@{}lrrrrrrr@{}}
\toprule
& \multicolumn{3}{c}{NLL by target labels} & \multicolumn{4}{c}{Overall} \\
\cmidrule(lr){2-4}\cmidrule(l){5-8}
Method & 16 & 32 & 64 & NLL $\downarrow$ & MAE $\downarrow$ & RMSE $\downarrow$ & Pred. $\rho\uparrow$ \\
\midrule
AR-Marginal & 2.2711 & 2.1052 & 1.9886 & 2.1216 & 23.8597 & \underline{35.3358} & \textbf{0.3191} \\
Support-CV & 2.3287 & 2.1309 & 1.9938 & 2.1511 & 26.2446 & 38.0759 & 0.3033 \\
Support-greedy & 2.3806 & 2.1695 & 2.0410 & 2.1970 & 28.6550 & 40.3969 & 0.2949 \\
Source-quality & 2.4467 & 2.2451 & 2.0806 & 2.2575 & 29.3549 & 41.0480 & 0.3111 \\
Residual & 2.5339 & 2.2696 & 2.0806 & 2.2947 & 32.3550 & 44.8218 & 0.3053 \\
Target-only & 2.4456 & 2.3084 & 2.2476 & 2.3339 & 29.3636 & 38.8208 & 0.1805 \\
AssayRouter-Z & 2.2666 & 2.0989 & 1.9889 & 2.1181 & 23.6652 & 35.3589 & \underline{0.3124} \\
AssayRouter-$\Delta$ & \underline{2.2647} & \underline{2.0926} & \textbf{1.9853} & \underline{2.1142} & \underline{23.5806} & 35.3864 & 0.2961 \\
AssayRouter-C & \textbf{2.2551} & \textbf{2.0896} & \underline{1.9860} & \textbf{2.1102} & \textbf{23.3872} & \textbf{35.0709} & 0.3011 \\
\bottomrule
\end{tabular}
}
\vspace{-4mm}
\end{table}

The overall strict NLL gap to Support-CV is 0.0409 (95\% CI $[0.0211,0.0623]$). AssayRouter-C has lower NLL in 77.8\% of target--interface units; in original units, mean MAE falls from 26.24 to 23.39, a 10.9\% reduction. Marginal-over-states attains the highest prediction Spearman correlation, distinguishing response ordering from numerical accuracy. Both exhaustive Support-CV and cheaper support-greedy remain weaker under strict scoring; support-greedy also trails at every budget and interface under prediction averaging (Appendix~\ref{app:matched-alternatives}).

\begin{figure}[ht]
\vspace{-1mm}
\centering
\includegraphics[width=0.98\textwidth]{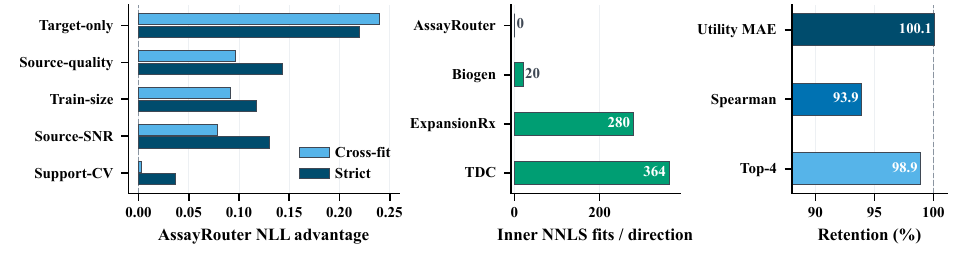}
\vspace{-5mm}
\caption{\textbf{Historical utility is specific and amortizes search on the target.} Left: matched controls against AssayRouter-$\Delta$ under cross-fit and strict scoring; positive values favor the historical prior. Middle: inner subset fits for Support-CV in each collection. AssayRouter performs no counterfactual NNLS fit on the target; both methods retain the same final NNLS combiner. Right: relative performance when both historical target and candidate source identities are held out; 100\% is the target-only out-of-fold reference, MAE is inverted, and values above 100\% indicate improvement over that reference.}
\vspace{-1mm}
\label{fig:acceptance-controls}
\end{figure}

Support-CV refits candidate subsets for every support split, whereas AssayRouter scores each source once and reuses the frozen ranking. The search requires 20, 280, and 364 inner NNLS fits per direction on Biogen, ExpansionRx, and TDC; AssayRouter performs none before the shared final combiner. Prediction averaging expands one episode to 64 source calls and more than 20 million inner fits across the evaluation. Strict scoring evaluates each four-call constituent independently. AssayRouter therefore improves the matched deployment action while amortizing its construction cost (Fig.~\ref{fig:acceptance-controls}, middle; Appendix~\ref{app:amortized-cost}).

The matched controls test the principal shortcuts when outputs are frozen. Target-only measures whether the bank contributes at all; source quality, training size, and support SNR replace learned utility with metadata or support statistics; residual compatibility uses the local target error profile; and Support-CV supplies subset search on the target. All retain the same Top-4 fan-in and final convex fit. The left panel of Fig.~\ref{fig:acceptance-controls} shows the resulting hierarchy: removing source reuse creates the largest gap, generic rules for source strength remain consistently weaker, and Support-CV approaches parity only when its many searched constituents are averaged. Strict scoring exposes the deployed four-call action and separates the frozen historical ranking from this search. Query-local DES and MINE, together with the all-source convex model, answer different questions about information access or fan-in (Appendix~\ref{app:reuse-paradigms}; Table~\ref{tab:set-conditional-main}).

\subsection{Post-fit supervision survives shortcut and stability tests}

Residual compatibility is the closest shortcut based on support profiles under the same four-source contract. AssayRouter-C improves strict NLL by 0.1844 over this selector (95\% CI $[0.1138,0.2609]$). The advantage appears in all six interfaces and at least 79\% of targets at every budget (Table~\ref{tab:strict-four-call-main}; Appendix~\ref{app:strict-residual-compatibility}). Residual correlation measures whether a source follows local-model errors; post-fit utility measures whether it remains useful after competing channels enter the normalized NNLS combiner. Their gap localizes the transferred information to constituent value inside the fitted ensemble. The candidate-label intervention, unseen-interface transfer, and strict Support-CV comparison also retain their direction after deleting any one target, interface, or collection (Appendix~\ref{app:small-cluster-sensitivity}).

\begin{figure}[ht]
\centering
\begin{minipage}[t]{0.66\textwidth}
\vspace{0pt}
\includegraphics[width=\linewidth]{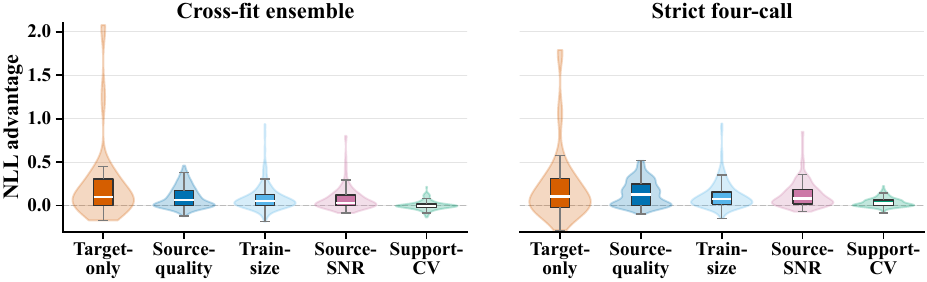}
\end{minipage}\hfill
\begin{minipage}[t]{0.30\textwidth}
\vspace{-8pt}
\caption{\textbf{Matched alternatives underperform.}
Distributions aggregate 144 target--interface NLL effects across budgets; positive
values favor AssayRouter-$\Delta$. Boxes mark quartiles and medians.}
\label{fig:matched-alternative-controls-intervals}
\end{minipage}
\end{figure}

Figure~\ref{fig:matched-alternative-controls-intervals} compares matched alternatives across target--interface units. Target-only and generic measures of source strength favor AssayRouter under both estimands. Support-CV reaches parity only through prediction averaging and favors AssayRouter when four-call actions are scored independently. Because all controls share assignments, Top-4 fan-in, and final NNLS, the hierarchy isolates each ranking signal from post-fit supervision. Appendix~\ref{app:matched-alternatives} gives the contracts and point effects.

\begin{figure}[ht]
\vspace{-1mm}
\centering
\includegraphics[width=0.98\textwidth]{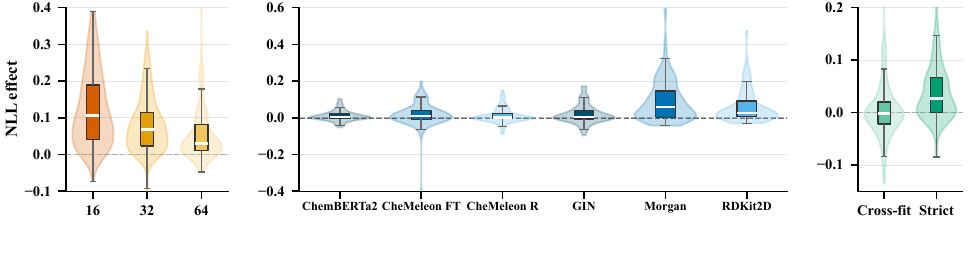}
\vspace{-8mm}
\caption{\textbf{The central effects span scientific units.}
Left: strict candidate-label intervention across 144 target--interface cells; middle:
unseen-interface transfer across 72 target--budget cells; right: Support-CV across 144
units under prediction averaging and strict scoring. Positive values favor
AssayRouter-$\Delta$; violins contain quartile boxes and medians.}
\label{fig:central-effect-distributions}
\vspace{-1mm}
\end{figure}

State features improve post-fit utility prediction without improving external routing (Fig.~\ref{fig:set-conditional-results}C). A refined score changes deployment only when it changes Top-4 membership and the ensuing refit changes the combination function. Utility-prediction MSE can therefore improve while preserving either the selected set or its refitted predictor, explaining why proxy regression and deployed decision quality separate. Figure~\ref{fig:central-effect-distributions} resolves the transferred object across scientific units: candidate value after fitting the local target model, expressed strongly enough to change a sparse action.

These distributions preserve assay heterogeneity: the strict intervention is broadest at 16 labels, unseen-interface effects are positive in aggregate for every held-out family, and strict Support-CV separates from prediction averaging under the four-call estimand. Agreement between the centered primary result (Table~\ref{tab:strict-four-call-main}) and uncentered $\Delta$ replication (Fig.~\ref{fig:central-effect-distributions}) shows that block centering does not create the effect. Historical ordering matters most before a new assay can search the bank reliably.

The deletion analysis comprises 24 leave-one-target, six leave-one-interface, and three leave-one-collection estimates per central comparison. Every deletion preserves the signs of the candidate-label intervention, unseen-interface effect, and strict Support-CV advantage; two-sided exact sign tests over the six interface effects give $p=0.03125$ for all three. Thus, no single assay, predictor family, or external collection drives the conclusions.

The support-scale sensitivity changes only the historical $Q_{\mathrm{disc}}$ loss scale while preserving routing on $R$, fitting on $C$, assignments, and deployment; direct, strict, unseen-interface, and strict search effects remain positive (Fig.~\ref{fig:set-conditional-results}D). The crossed identity audit predicts each historical row without the identity of its pseudo-target or candidate source, yet utility MAE and Top-4 overlap match the reference that holds out the target (Fig.~\ref{fig:acceptance-controls}, right). Together with leave-one-interface-out deployment, these results identify a relation from candidate behavior to utility that is shared across assays and prediction contracts.

\section{Related Work}

\noindent\textbf{Molecular adaptation and representation.} ActFound uses pairwise meta-learning~\citep{feng2024actfound}; \mbox{FS-CAP}~\citep{eckmann2024fscap} and UniMatch~\citep{li2025unimatch} condition on labeled sets; \mbox{MolSetRep} learns set representations~\citep{boulougouri2024molsetrep}; and Contrastive KERMT transfers pretrained ADME representations~\citep{xue2026contrastivekermt}. Other approaches span multitask benchmarks~\citep{stanley2021fsmol}, adaptive kernels~\citep{chen2023adkf}, contextual memory~\citep{schimunek2023mhnfs}, in-context prediction~\citep{fifty2023camp}, and learned task transfer~\citep{lee2024gate,zhong2025autaut}. These methods train, fine-tune, or condition a predictor. AssayRouter keeps source contracts frozen and learns which outputs complement a local model. Appendix~\ref{app:external-reproductions} reproduces official systems; access-matched routing controls form the main comparison.

\noindent\textbf{Source and task selection.} Prediction-profile methods use statistical relations between source outputs and a target: pQSAR transfers dense cross-assay correlations~\citep{martin2019pqsar,martin2021collaborative}, while MINE estimates query-local competence~\citep{moura2021mine}. Task-level or auxiliary evidence guides transfer weights in GATE~\citep{lee2024gate}, attribution-supervised assay selection in AssayMatch~\citep{fan2026assaymatch}, and imputation in QComp~\citep{yang2025qcomp}. TaskWeb~\citep{kim2023taskweb} and MetaGL~\citep{park2023metagl} likewise transfer task or model relations. These signals describe relatedness or competence before combination; AssayRouter supervises value after the local predictor and source output are fitted together.

\noindent\textbf{Model portfolios and learned routing.} Ensemble selection ranks validation libraries~\citep{caruana2004ensemble}; zero-shot AutoML ranks pipelines~\citep{ozturk2022zero}; Model Spider ranks and adapts one model from a pretrained zoo~\citep{zhang2023modelspider}; limited-label selection chooses one pretrained classifier~\citep{okanovic2025modelselector}; vision-language reuse selects several models per class~\citep{tan2025mll}; and RouteLLM allocates one frozen language model~\citep{ong2025routellm}. Black-box meta-learning transfers through inaccessible APIs~\citep{hu2023apis}. AssayRouter selects a fixed-size subset, freezes membership before confirmation, and fits convex weights afterward. Post-fit utility supervision is the contribution; HGB supplies the estimator. Matched comparators retain the bank, $R/C$ roles, fan-in, and final NNLS, while other actions and information budgets provide positioning precedents (Appendix~\ref{app:reuse-paradigms}; Table~\ref{tab:baseline-coverage}).

\section{Discussion and Conclusion}

Historical supervision from completed ChEMBL-MT assays transfers across three external collections and six frozen interfaces. Candidate-label interventions establish that utility must remain paired with candidate behavior; identity and interface holdouts show that this relation extends beyond memorized assays or predictor families. The advantage is strongest at 16 labels and narrows as evidence on the target grows, identifying the scarce-label regime in which amortized historical routing matters most.

The separation between historical utility prediction and external NLL gives a broader evaluation principle. A better proxy matters only when it changes membership near the Top-4 boundary and the refitted combiner assigns consequential weight to that change. Router selection should therefore validate the deployed decision under its exact information and call budget, rather than rely on proxy accuracy alone.

The demonstrated scope is continuous regression with homogeneous interface banks, a fixed four-source action, and 20 historical pseudo-targets. Fixing $K=4$ makes every selector answer the same sparse decision; learning target-dependent fan-in would extend the action space. Expanding the historical panel would test how utility transfer scales with assay diversity, while learned interface representations could replace the current family code when a deployed contract has no historical counterpart. Classification endpoints require a corresponding post-fit utility loss.

Post-fit routing can also complement few-shot adaptation: AssayRouter selects the frozen channel set, after which an adaptive method can operate within that set. The present evidence establishes this selection step. Completed assays provide reusable supervision for selecting frozen prediction channels, improving strict routing while removing repeated subset search on the target.

\bibliography{references}
\bibliographystyle{iclr2027_conference}

\appendix

\section{Protocol and Decision Controls}
\label{app:set-conditional-protocol}

This appendix specifies the training protocol and maps each central claim to its matched control and statistical unit. Tables~\ref{tab:information-budget-main}--\ref{tab:central-evidence-chain} define access, alternative explanations, and the central evidence chain; Tables~\ref{tab:qdisc-scale-sensitivity}--\ref{tab:compound-nonoverlap} report the interventions and sensitivity analyses; Appendix~\ref{app:routing-comparators} contains the matched comparators; and Appendix~\ref{app:amortized-cost} analyzes selection cost. Appendix~\ref{app:retrospective-development} records the development boundary, and Appendix~\ref{app:external-reproductions} separates reproductions that use full source training from the routing comparison at matched support budgets.

\subsection{Training, interventions, and audits}
\label{app:training-audits}

\subsubsection{Historical matrix}
\label{app:historical-matrix}
AssayRouter is trained only on continuous ChEMBL-MT~\citep{adrian2025chemblmt} pseudo-targets. Each of 20 targets is crossed with six predictor interfaces, support budgets 16, 32, and 64, and four historical episodes. Each support is divided deterministically into $R$ and $C$; a third role $Q_{\mathrm{disc}}$ is disjoint from both. For support budget $n$, the two fitting roles contain $n/2$ molecules each (8, 16, or 32), while $Q_{\mathrm{disc}}$ draws up to 128 molecules from an independent internal discovery split (25--128 across the 20 historical assays). The primary target is the reduction in discovery loss from adding one candidate to the local target predictor. This gives 27,360 distinct candidate contexts. To match the refinements' state frequency, each candidate context is repeated over the same grid of 13 states, yielding 355,680 rows. The router is a histogram gradient-boosting regressor conditioned on the interface and selected by grouped cross-validation over target identity.

\subsubsection{Frozen fitting configuration}
\label{app:frozen-fitting}
The local target learner is Ridge regression with $\alpha=1$ on molecular fingerprints. Four shuffled folds produce its out-of-fold support column; the fit on all $C$ rows produces query predictions. Regression labels are centered by the median and divided by $1.4826$ times the median absolute deviation, falling back to the sample standard deviation and then $10^{-6}$. The local column and selected source columns are fitted by NNLS and normalized to the simplex. HGB model selection compares three fixed configurations $(\eta,T,L,m,\lambda)$: $(0.05,200,15,20,1)$, $(0.05,300,31,30,1)$, and $(0.08,220,31,40,3)$, denoting learning rate, iterations, maximum leaves, minimum leaf samples, and $\ell_2$ regularization. Early stopping is disabled. Five folds grouped by target select MAE averaged over targets, after which the winning configuration is refitted on the complete historical matrix.

\begin{table}[ht]
\centering
\setlength{\tabcolsep}{5pt}
\caption{Information available to the principal selectors. Query-local methods may
change the selected set across confirmation molecules.}
\label{tab:information-budget-main}
\begin{tabular}{@{}lccc@{}}
\toprule
Selector & History & Full-bank query & Query-local \\
\midrule
AssayRouter & Utility & No & No \\
Support-CV & None & No & No \\
Frozen-DES & None & Yes & Yes \\
MINE-WS~\citep{moura2021mine} & Competence & Yes & Yes \\
All-source & None & Yes & No \\
\bottomrule
\end{tabular}
\end{table}

\begin{table}[ht]
\centering
\small
\setlength{\tabcolsep}{3.2pt}
\caption{Coverage of alternative explanations relevant to the decision across five contract
axes. Each control changes one scientific axis while preserving the frozen prediction
contract wherever that axis permits.}
\label{tab:baseline-coverage}
\begin{tabular}{@{}lll@{}}
\toprule
Alternative explanation & Matched test & Changed axis \\
\midrule
No value from bank reuse & Target-only & Bank reuse \\
Generic source strength suffices & Source-quality & Utility supervision \\
Metadata/support statistics suffice & Size / SNR & Ranking signal \\
Profile compatibility suffices & Residual (strict) & Ranking signal \\
Candidate--label pairing is arbitrary & Frozen permutations & Ranking signal \\
Target-side search suffices & Support-CV@4 & Access/timing \\
Query-local competence suffices & DES@4 / MINE-WS@4 & Access/timing \\
Sparse fan-in creates the gain & All-source convex & Action/fan-in \\
Calibration or state creates the gain & $\Delta$ / C / Z / Marginal / State & Utility supervision \\
\bottomrule
\end{tabular}
\end{table}

\begin{table}[ht]
\centering
\caption{Evidence-chain index. Effects favor AssayRouter-C; the final column reports
the scientific unit appropriate to each test.}
\label{tab:central-evidence-chain}
\begin{tabular}{@{}lrl@{}}
\toprule
Test & NLL effect & Positive breadth \\
\midrule
Candidate-label intervention & 0.1352 & 90.7\% of cells \\
Unseen-interface intervention & 0.0497 & 6/6 held-out families \\
Support-CV@4 vs. C & 0.0409 & 77.8\% of target--interface units \\
Residual compatibility vs. C & 0.1844 & 83.3\% of targets \\
\bottomrule
\end{tabular}
\end{table}

Source-quality is the capacity-matched learned selector: it retains all 11 candidate coordinates, HGB capacity search, grouped folds, the matched expansion over 13 states, Top-4 action, external assignments, and final NNLS, changing only the historical supervision label. It therefore tests whether learning a selector from generic source strength can explain the result. Additional model selection methods with the same inputs and action would repeat this scientific role; a new numerical comparator is relevant only when it instantiates an uncovered supervision, access, timing, or action axis.

\subsubsection{Support-scale $Q_{\mathrm{disc}}$ utility-loss sensitivity}
\label{app:qdisc-scale-sensitivity}
For each frozen historical episode, we recompute only the utility loss as $\ell_{\mathrm{NLL}}(e_C\sigma_C/\sigma_{R\cup C})$, where $e_C$ is the original discovery residual standardized on $C$. Routing features remain local to $R$, the local target and normalized NNLS fits remain local to $C$, and external deployment is unchanged. The sensitivity reuses the original direct and leave-one-interface-out donor mappings and every external assignment. Each paired decision uses the same 24 targets, six interfaces, three budgets, and 12 support episodes as its parent experiment, with the target--interface bootstrap of 10,000 resamples defined in Appendix~\ref{app:set-statistics}. The prespecified criteria require positive NLL interval lower bounds for the direct, strict, unseen-interface, and strict-search comparisons, plus positive unseen-interface effects in at least four of six held-out families. The direct, strict, unseen-interface, and matched-search comparisons preserve the candidate-specific utility advantage (Table~\ref{tab:qdisc-scale-sensitivity}).

\begin{table}[ht]
\centering
\caption{Support-scale $Q_{\mathrm{disc}}$ sensitivity. Values are paired point
effects; all five prespecified criteria are met.}
\label{tab:qdisc-scale-sensitivity}
\begin{tabular}{@{}lcc@{}}
\toprule
Comparison & Effect & Value \\
\midrule
Direct intervention & Permuted $-$ genuine NLL & $+0.0463$ \\
Strict intervention & Permuted $-$ genuine NLL & $+0.0816$ \\
Unseen interface & Permuted $-$ genuine NLL & $+0.0299$ \\
Unseen interfaces & Positive held-out families & $5/6$ \\
Strict search & Support-CV $-$ genuine NLL & $+0.0439$ \\
\bottomrule
\end{tabular}
\end{table}

\subsubsection{Source-identity holdout}
\label{app:source-identity}
The 20 historical candidate-source identities are disjoint from the 37 source-assay identities in the external prediction banks. We additionally cross-fit the historical matrix over five target folds and five candidate-source folds, training each of 25 HGB models only on rows whose pseudo-target and candidate source both lie outside the held folds. Every one of the 355,680 historical rows is predicted exactly once, with source identity absent from the features. Relative to the original target-held-out predictions, the double-held-out model has matched utility MAE ($-0.00006$, 95\% interval $[-0.00184,0.00178]$) and matched Top-4 overlap ($-0.00382$, $[-0.02205,0.01493]$); fine-grained Spearman correlation decreases by $0.0197$ ($[-0.0355,-0.0025]$). The deployment candidate set therefore transfers across historical source identities; the observed decrease is confined to fine-grained ordering beyond the selection boundary.

\subsubsection{Matched post-fit label controls}
\label{app:postfit-label-controls}
These controls retain the historical rows, $R/C/Q_{\mathrm{disc}}$ roles, routing features, interface code, grouped HGB search, external assignments, Top-$4$ rule, and simplex fit. They replace only the historical label with negative post-fit loss after adding the candidate, its percentile rank within context, its value centered within the block, or its value standardized within the block. Raw loss and rank are weaker than standalone utility. AssayRouter-C and AssayRouter-Z remain matched to AssayRouter-$\Delta$ in overall strict NLL; their control-minus-$\Delta$ effects are $-0.0039$ with interval $[-0.0102,0.0017]$ and $+0.0039$ with interval $[-0.0044,0.0130]$. This identifies post-fit candidate value relative to its historical block as the transferable supervision shared by the successful constructions.

\begin{table}[ht]
\centering
\caption{Strict matched post-fit controls. Point estimates are averaged after each
four-call constituent is scored.}
\label{tab:postfit-label-controls}
\begin{tabular}{@{}lrrrr@{}}
\toprule
Method & NLL $\downarrow$ & MAE $\downarrow$ & RMSE $\downarrow$ & Spearman $\uparrow$ \\
\midrule
AssayRouter-$\Delta$ & 2.1142 & 23.5806 & 35.3864 & 0.2961 \\
AssayRouter-C & 2.1102 & 23.3872 & 35.0709 & 0.3011 \\
AssayRouter-Z & 2.1181 & 23.6652 & 35.3589 & 0.3124 \\
Post-fit loss & 2.1684 & 25.9818 & 37.2848 & 0.3002 \\
Post-fit rank & 2.1515 & 24.9610 & 36.3301 & 0.3228 \\
\bottomrule
\end{tabular}
\end{table}

\subsubsection{Direct AssayRouter-C correspondence intervention}
\label{app:direct-c-intervention}
The primary centered label receives the same frozen candidate bijection used by the standalone-utility intervention. Permutation and block centering commute to numerical precision, so the manipulation preserves every block label multiset while changing only which candidate receives each value. The same HGB search, external assignments, Top-$4$ rule, and simplex fit are replayed. Table~\ref{tab:block-centered-permutation} shows that the candidate--label pairing remains decisive for the selected deployment variant under both estimands.

\begin{table}[h]
\centering
\caption{Direct AssayRouter-C candidate-label intervention. NLL effects are permuted
minus genuine; win rate is the fraction of 432 target--interface--budget cells favoring
the genuine mapping. Intervals use target--interface clustered resampling.}
\label{tab:block-centered-permutation}
\begin{tabular}{@{}lrrr@{}}
\toprule
Estimand & NLL effect & 95\% interval & Cell win rate \\
\midrule
Cross-fit & 0.0814 & $[0.0427,0.1246]$ & 0.7222 \\
Strict & 0.1352 & $[0.0852,0.1915]$ & 0.9074 \\
\bottomrule
\end{tabular}
\end{table}

Tables~\ref{tab:evaluation-matrix} and~\ref{tab:interface-bank-main} give the complete training/evaluation inventory and the six fixed interface contracts used by the reported AssayRouter experiments.

\begin{table}[h]
\centering
\caption{Training and frozen external evaluation. External target identities and
confirmation labels are absent from router training and model selection.}
\label{tab:evaluation-matrix}
\begin{tabular}{@{}llll@{}}
\toprule
Role & Collection & Targets & Interfaces \\
\midrule
Historical supervision & ChEMBL-MT~\citep{adrian2025chemblmt} & 20 & 6 \\
Temporal holdout & ExpansionRx~\citep{openadmet2026expansionrx} & 9 & 6 \\
Sparse-panel holdout & Biogen~\citep{fang2023prospective} & 6 & 6 \\
Scaffold holdout & TDC regression~\citep{huang2021tdc} & 9 & 6 \\
\bottomrule
\end{tabular}
\end{table}

\begin{table}[h]
\centering
\caption{The six formal interfaces to frozen predictors. Abbreviations keep each cell on
one line: FP, emb., repr., and GIN denote fingerprint, embedding, representation, and
graph-isomorphism network. A bank contains independently trained source-assay
predictors from one interface family.}
\label{tab:interface-bank-main}
\begin{tabular}{@{}lll@{}}
\toprule
Molecular representation & Source predictor & Reuse mode \\
\midrule
Morgan FP~\citep{rogers2010extended} & Ridge & Contract \\
RDKit2D~\citep{rdkitsoftware} & LightGBM~\citep{ke2017lightgbm} & Contract \\
CheMeleon emb.~\citep{burns2025descriptor} & Ridge & Frozen repr. \\
ChemBERTa2 emb.~\citep{ahmad2022chemberta2} & LightGBM~\citep{ke2017lightgbm} & Frozen repr. \\
Molecular graph & GIN~\citep{xu2019gin} & Per-source \\
CheMeleon repr.~\citep{burns2025descriptor} & Prediction head & Fine-tuned \\
\bottomrule
\end{tabular}
\end{table}

Within each source, Morgan/Ridge and CheMeleon/Ridge use 12 fixed members, RDKit2D/LightGBM and ChemBERTa2/LightGBM use eight, and GIN and the fine-tuned CheMeleon head use three. The interface returns the member mean as $z_s(x)$ and their empirical standard deviation as $u_s(x)$.

The feature inventory below separates the primary 11-coordinate AssayRouter input from the 14 state coordinates used only by the extension.

\subsubsection{Feature inventory}
\label{app:feature-inventory}
\label{tab:set-features}
The primary 11-coordinate HGB utility prior uses signed and absolute residual alignment, residual correlation, source mean and standard deviation, absolute source response, uncertainty mean and spread, signal-to-noise ratio, log size of the routing support, and declared training size. The state-conditioned extension adds signed and absolute remaining alignment, remaining correlation, mean and maximum redundancy among selected sources, set size, residual scale, variance reduction, sum of source weights, number of active sources, current and augmented condition numbers, and effective rank. The six interface indicators are appended separately to every design. All coordinates are available before confirmation labels are read; coordinates derived from predictions use $R$, and training size is interface metadata.

The three routers share the 11 coordinates for each candidate used by AssayRouter and both refinements. The remaining 14 coordinates are available only to the state-conditioned extension. The primary router represents each candidate context by 13 equally weighted copies with identical candidate features and standalone label. Marginal-over-states uses the 13 labels specific to each state, while the state-conditioned extension additionally exposes the partial set. This schema of 11 candidate and 14 state coordinates is separate from the earlier pairwise feature design summarized in Appendix~\ref{app:retrospective-development}.

All three regressors append the same interface code to those coordinates. The full historical model uses six one-hot indicators. Each leave-one-interface-out model fits an encoder over the five observed interfaces and maps the held-out family to a five-dimensional zero vector, with no calibration specific to that interface.

Table~\ref{tab:standalone-hur-intervals} compares standalone and marginal-over-states supervision with the same estimator, rows, and external replay.

\begin{table}[h]
\centering
\caption{Matched supervision refinements. Differences are marginal-over-states utility
NLL minus standalone-utility NLL; positive values favor the primary utility prior.}
\label{tab:standalone-hur-intervals}
\begin{tabular}{lr}
\toprule
Scope & Mean difference \\
\midrule
Overall & $-0.0038$ \\
16 labels & $-0.0067$ \\
32 labels & $-0.0002$ \\
64 labels & $-0.0043$ \\
Biogen & $-0.0035$ \\
ExpansionRx & 0.0008 \\
TDC & $-0.0085$ \\
\bottomrule
\end{tabular}
\end{table}

Table~\ref{tab:refinement-ladder} aligns the supervision, representation, and external decision quality of the three nested routers.

\begin{table}[h]
\centering
\caption{Prediction refinement and external decision quality. Historical out-of-fold
(OOF) mean-squared error (MSE) is defined only for the shared marginal-over-states
label; lower is better.}
\label{tab:refinement-ladder}
\begin{tabular}{lccr}
\toprule
Router & Historical target & State features & External NLL \\
\midrule
AssayRouter (standalone-utility prior) & $\Delta(s\mid\varnothing)$ & No & 2.0418 \\
Marginal-over-states utility & $\mathbb E_A\Delta(s\mid A)$ & No & 2.0380 \\
State-conditioned extension & $\Delta(s\mid A)$ & Yes & 2.0392 \\
\bottomrule
\end{tabular}
\end{table}

Table~\ref{tab:nested-utility-oof} then isolates prediction refinement on the shared marginal target before any external routing decision is made.

\begin{table}[h]
\centering
\caption{Nested post-fit utility-prediction audit. Both columns use frozen
target-grouped out-of-fold predictions for the same marginal labels.}
\label{tab:nested-utility-oof}
\begin{tabular}{lrrr}
\toprule
Scope & Marginal utility MSE & State-conditioned MSE & Relative reduction \\
\midrule
Overall & 0.02064 & 0.01949 & 5.58\% \\
16 labels & 0.02584 & 0.02419 & 6.41\% \\
32 labels & 0.02061 & 0.01923 & 6.72\% \\
64 labels & 0.01546 & 0.01504 & 2.66\% \\
\bottomrule
\end{tabular}
\end{table}

\subsubsection{Historical-label permutations}
\label{app:historical-permutations}
The primary intervention uses one prespecified bijection, independent of labels, to permute standalone utility among 19 candidates inside each block defined by target, interface, support budget, and episode. Candidate features and the label multiset remain fixed, and the same permuted label is repeated over the matched grid of 13 states. The frozen mapping moves candidates in every block and retains 5.32\% fixed points overall. The clustered interval estimates the external effect of this frozen intervention; no Monte Carlo permutation $p$-value is inferred. A supporting intervention permutes the \emph{complete utility trajectory over 13 states} for marginal-over-states utility. It preserves all 355,680 labels, coverage of set sizes, and trajectory structure. Both interventions retain estimator capacity, grouped validation, external assignments, $K=4$ combination, and the boundary on confirmation labels.

The main bijection is replicate 1 in Figure~\ref{fig:set-conditional-results}B. Four additional fixed mappings repeat the intervention without changing the training or replay protocol. Table~\ref{tab:multi-permutation-stability} reports their external point effects; every replicate has a positive clustered lower bound.

\begin{table}[h]
\centering
\caption{Independent candidate-label interventions under the cross-fit prediction-average
estimand. Each row uses a distinct frozen bijection; effects are permuted minus genuine NLL.}
\label{tab:multi-permutation-stability}
\begin{tabular}{lrr}
\toprule
Frozen mapping & Mean NLL effect & Cell win rate \\
\midrule
1 & 0.0494 & 73.38\% \\
2 & 0.0477 & 70.83\% \\
3 & 0.0146 & 60.65\% \\
4 & 0.0339 & 72.69\% \\
5 & 0.0454 & 71.53\% \\
\bottomrule
\end{tabular}
\end{table}

\subsubsection{External replay}
\label{app:external-replay}
The three external collections by six interfaces form 18 evaluated deployments. Each target--interface--budget cell has 12 support episodes. Every episode produces eight balanced $R/C$ partitions and two role directions. The prediction-averaged estimand averages the 16 raw-scale predictions before scoring; the strict estimand scores each four-source constituent and then averages its scalar losses. Sparse methods select exactly four source inputs per constituent, and all methods fit the same no-prior simplex NNLS on $C$. The target predictor is deleted from the bank, external targets never enter router training, and confirmation labels enter only after all predictions have been frozen.

\subsubsection{Statistical unit and direction}
\label{app:set-statistics}
Support repetitions, partitions, and role directions are averaged before scientific comparison. A cell is a collection--target--interface--budget combination. Aggregate uncertainty uses a two-way pigeonhole bootstrap that independently resamples target and interface clusters 10,000 times. Unless stated otherwise, positive paired differences mean comparator NLL minus the stated AssayRouter variant.

\subsubsection{Small-cluster sensitivity}
\label{app:small-cluster-sensitivity}
We recompute the three headline NLL effects after leaving out each target, interface, or collection in turn, with frozen predictions and no model refitting. The strict candidate-label intervention, leave-one-interface-out transfer, and strict Support-CV comparison remain positive in every deletion. Exact sign-flip tests over the six interface effects give $p=0.03125$ for all three comparisons; sign-flip tests over the 24 target effects also remain positive. This sensitivity treats the small number of clusters directly and agrees with the target--interface bootstrap used for the primary analysis.

\subsubsection{Evidence design}
\label{app:evidence-design}
Five controlled changes answer distinct questions while preserving frozen assignments and the boundary on confirmation labels. Candidate-label permutation tests whether post-fit utility remains informative when paired with the correct source; leave-one-interface-out training tests transfer to an unseen family; source-quality supervision tests whether generic source strength suffices; the target-only learner tests whether the frozen bank adds value; and strict constituent scoring tests whether the result survives the deployed call budget.

\subsubsection{Matched alternative explanations}
\label{app:matched-alternatives}
The target-only control keeps the same cross-fitted local learner, support roles, and confirmation scoring while deleting every source call. The source-quality prior keeps the complete AssayRouter training and replay pipeline but replaces post-fit utility with the improvement on the discovery set of an affine-calibrated source over a constant predictor; its label never fits or evaluates the local target model. Two deterministic rankings select the four sources with the largest declared training sets or the strongest signal-to-noise ratio on the routing set. Support-greedy uses routing-support labels to add one source at a time, then freezes the set before the same final fit and confirmation scoring. These controls share the assignments and are evaluated under both prediction-averaged and strict constituent estimands. Table~\ref{tab:matched-alternative-controls} reports point effects; Fig.~\ref{fig:matched-alternative-controls-intervals} shows their target--interface distributions.

\begin{table}[h]
\centering
\caption{\textbf{Matched alternatives.} Paired mean NLL differences; positive values favor AssayRouter-$\Delta$.}
\label{tab:matched-alternative-controls}
\begin{tabular}{@{}lrr@{}}
\toprule
Alternative & Cross-fit & Strict \\
\midrule
Target-only & 0.2403 & 0.2197 \\
Source-quality & 0.0969 & 0.1433 \\
Train-size & 0.0912 & 0.1174 \\
Source-SNR & 0.0789 & 0.1304 \\
Support-CV & 0.0030 & 0.0369 \\
\bottomrule
\end{tabular}
\end{table}

Table~\ref{tab:set-conditional-main} preserves the exact values behind Figure~\ref{fig:set-conditional-results}A. Rows are grouped by scientific role because refinements, information-rich controls, interventions, and classical diagnostics do not form a single access-matched leaderboard.

\begin{table}[h]
\centering
\caption{\textbf{Stability-focused cross-fit estimand.} Absolute
support-standardized robust NLL on 24 external targets excluded from router fitting and six prediction interfaces;
lower is better. All sparse constituent models have four-source combiner fan-in, but
their information permissions differ as described in the text.}
\label{tab:set-conditional-main}
\begin{tabular}{@{}lrrr@{}}
\toprule
Method & 16 labels & 32 labels & 64 labels \\
\midrule
AssayRouter (standalone-utility prior) & 2.1606 & 2.0300 & 1.9348 \\
\addlinespace[2pt]
\multicolumn{4}{@{}l}{\textit{Matched refinements}} \\
Marginal-over-states utility & 2.1539 & 2.0297 & 1.9305 \\
State-conditioned extension & 2.1589 & 2.0285 & 1.9301 \\
\addlinespace[2pt]
\multicolumn{4}{@{}l}{\textit{Search and information-rich controls}} \\
Support-CV@4 ensemble & 2.1806 & 2.0315 & 1.9223 \\
Frozen-DES@4 & 2.1678 & 2.0268 & 1.9231 \\
MINE-WS@4~\citep{moura2021mine} & 2.2058 & 2.0611 & 1.9309 \\
All-source convex & 2.1771 & 2.0351 & 1.9303 \\
\addlinespace[2pt]
\multicolumn{4}{@{}l}{\textit{Mechanism intervention and classical diagnostics}} \\
AssayRouter, utility labels permuted & 2.2419 & 2.0714 & 1.9602 \\
Residual compatibility & 2.3634 & 2.1490 & 1.9949 \\
Support-greedy selection~\citep{caruana2004ensemble} & 2.2133 & 2.0530 & 1.9572 \\
50 random four-source sets & 2.2212 & 2.0619 & 1.9515 \\
\bottomrule
\end{tabular}
\end{table}

Table~\ref{tab:standalone-permutation-intervals} reports the direct intervention at the aggregate, budget, collection, and interface levels.

\begin{table}[h]
\centering
\caption{Direct standalone-utility permutation. Differences are permuted-router NLL
minus genuine AssayRouter NLL; positive values favor genuine post-fit labels.}
\label{tab:standalone-permutation-intervals}
\begin{tabular}{lrr}
\toprule
Scope & Mean difference & Win rate \\
\midrule
Overall & 0.0494 & 73.38\% \\
16 labels & 0.0813 & 84.03\% \\
32 labels & 0.0415 & 72.22\% \\
64 labels & 0.0254 & 63.89\% \\
Biogen & 0.0030 & 50.00\% \\
ExpansionRx & 0.0865 & 85.19\% \\
TDC & 0.0433 & 77.16\% \\
\midrule
ChemBERTa2 / LightGBM & 0.0838 & 80.56\% \\
Fine-tuned CheMeleon & 0.0321 & 68.06\% \\
Frozen CheMeleon / Ridge & 0.0397 & 66.67\% \\
GIN from scratch & 0.0316 & 70.83\% \\
Morgan / Ridge & 0.0327 & 75.00\% \\
RDKit2D / LightGBM & 0.0766 & 79.17\% \\
\bottomrule
\end{tabular}
\end{table}

The direct intervention moves all four matched metrics in the same scientific direction. Table~\ref{tab:standalone-permutation-secondary} keeps errors in the original units separate from support-standardized NLL and reverses the Spearman subtraction so that positive entries always favor genuine post-fit labels.

\begin{table}[h]
\centering
\caption{Secondary metrics for the direct standalone-utility permutation. NLL, MAE,
and RMSE are permuted minus genuine; Spearman is genuine minus permuted.}
\label{tab:standalone-permutation-secondary}
\begin{tabular}{lr}
\toprule
Metric & Mean effect \\
\midrule
Robust NLL & 0.0494 \\
MAE & 3.5805 \\
RMSE & 3.1850 \\
Spearman correlation & 0.0327 \\
\bottomrule
\end{tabular}
\end{table}

Tables~\ref{tab:strict-permutation-nll} and~\ref{tab:strict-permutation-secondary} replay the prespecified primary mapping under the strict deployment estimand. Each constituent is scored before aggregation, so every confirmation prediction uses exactly four source calls and no averaged prediction vector.

\begin{table}[h]
\centering
\caption{Strict four-call standalone-utility intervention. Differences are
permuted minus genuine NLL; positive values favor AssayRouter.}
\label{tab:strict-permutation-nll}
\begin{tabular}{lr}
\toprule
Scope & Mean difference \\
\midrule
Overall & 0.0870 \\
16 labels & 0.1260 \\
32 labels & 0.0816 \\
64 labels & 0.0533 \\
Biogen & 0.0263 \\
ExpansionRx & 0.1081 \\
TDC & 0.1063 \\
\midrule
ChemBERTa2 / LightGBM & 0.1295 \\
Fine-tuned CheMeleon & 0.0681 \\
Frozen CheMeleon / Ridge & 0.0940 \\
GIN from scratch & 0.0494 \\
Morgan / Ridge & 0.0675 \\
RDKit2D / LightGBM & 0.1133 \\
\bottomrule
\end{tabular}
\end{table}

\begin{table}[h]
\centering
\caption{Overall secondary effects for the strict four-call intervention. Losses are
permuted minus genuine; Spearman is genuine minus permuted.}
\label{tab:strict-permutation-secondary}
\begin{tabular}{lr}
\toprule
Metric & Mean effect \\
\midrule
Robust NLL & 0.0870 \\
MAE & 4.6262 \\
RMSE & 4.9810 \\
Spearman correlation & 0.0405 \\
\bottomrule
\end{tabular}
\end{table}

Table~\ref{tab:loio-permutation} asks whether the primary candidate--utility mapping centered within each block transfers when an entire predictor family is absent from historical supervision. Each router trains on the other five families, maps the held-out interface to a five-dimensional zero vector, forbids calibration specific to that interface, and reuses the same 5,184 external episode pairs.

\begin{table}[h]
\centering
\caption{AssayRouter-C leave-one-interface-out evaluation. Differences are
comparator minus genuine-router NLL after the named family is removed from historical
supervision; positive values favor AssayRouter-C.}
\label{tab:loio-permutation}
\begin{tabular}{lrr}
\toprule
Held-out scope & Permuted & Support-CV \\
\midrule
Overall & 0.0497 & 0.1116 \\
16 labels & 0.0697 & 0.1769 \\
32 labels & 0.0492 & 0.1029 \\
64 labels & 0.0301 & 0.0550 \\
Biogen & $-0.0023$ & 0.0367 \\
ExpansionRx & 0.0770 & 0.1594 \\
TDC & 0.0570 & 0.1138 \\
\midrule
ChemBERTa2 / LightGBM & 0.0060 & 0.1117 \\
Fine-tuned CheMeleon & 0.0188 & 0.0908 \\
Frozen CheMeleon / Ridge & 0.0121 & 0.1228 \\
GIN from scratch & 0.0673 & 0.0955 \\
Morgan / Ridge & 0.0963 & 0.1298 \\
RDKit2D / LightGBM & 0.0975 & 0.1190 \\
\bottomrule
\end{tabular}
\end{table}

\subsubsection{Compound provenance and nonoverlap sensitivity}
\label{app:compound-provenance}
Canonical parents are the largest sanitized fragments represented by canonical isomeric SMILES. The analysis detects no leakage of target labels. Biogen and ExpansionRx have zero exact-parent overlap with training data for the historical router or source predictors; TDC contains compound exposure across assays. For TDC, $H$-clean removes confirmation parents used to train the historical router, $S$-clean removes parents used to train any other source predictor, and Union-clean applies both filters. Stored assignments, selections, weights, and predictions are rescored without retraining or reselection. All six interface replays reproduce metrics on the full set to numerical precision before filtering, and confirmation labels are read only after cell predictions are frozen. A target enters clustered inference only when at least 16 confirmation molecules remain; the clearance microsome AZ endpoint retains five Union-clean rows and is excluded, leaving eight TDC targets. This sensitivity measures exact-parent overlap; scaffold independence lies outside its scope. Table~\ref{tab:compound-nonoverlap} reports each frozen filtering scope.

\begin{table}[h]
\centering
\caption{TDC exact-parent nonoverlap sensitivity. Differences are permuted minus
genuine NLL for the AssayRouter-$\Delta$ replication; positive values favor historical
candidate--utility correspondence. Retention is the mean confirmation fraction.}
\label{tab:compound-nonoverlap}
\begin{tabular}{llrrr}
\toprule
Router & Scope & Targets & Retention & Mean difference \\
\midrule
Direct & Full & 9 & 100.00\% & 0.0433 \\
Direct & $H$-clean & 9 & 62.38\% & 0.0419 \\
Direct & $S$-clean & 8 & 55.97\% & 0.0403 \\
Direct & Union-clean & 8 & 42.71\% & 0.0385 \\
\midrule
LOIO & Full & 9 & 100.00\% & 0.0525 \\
LOIO & $H$-clean & 9 & 62.38\% & 0.0512 \\
LOIO & $S$-clean & 8 & 55.97\% & 0.0464 \\
LOIO & Union-clean & 8 & 42.71\% & 0.0468 \\
\bottomrule
\end{tabular}
\end{table}

\subsection{Routing comparators and search cost}
\label{app:routing-comparators}

The following inventory separates AssayRouter from the closest paradigms for reusing models and tasks by the supervision available before deployment and the action taken on a new task.

\subsubsection{Closest reuse paradigms}
\label{app:reuse-paradigms}
\label{tab:novelty-boundary}
All-Assay-Max2 and collaborative pQSAR use prediction profiles and target--assay correlation to filter channels and fit target PLS without a fixed $K$ \citeyearpar{martin2019pqsar,martin2021collaborative}. Model Label Learning uses capability labels for each concept to select top-$k$ VLMs per class and ensemble zero-shot predictions \citeyearpar{tan2025mll}. Model Spider uses approximate historical model rankings to rank a model zoo and adapt its highest-ranked model \citeyearpar{zhang2023modelspider}. AssayMatch uses assay compatibility supervised by TRAK to rank assays for an unlabeled target and train on selected data \citeyearpar{fan2026assaymatch}. AssayRouter uses standalone post-fit loss reduction to select one global Top-4 set from the support data and then fits a convex combiner.

The cross-fit Support-CV comparison uses AssayRouter-$\Delta$, whose utility is uncentered; the primary strict comparison uses AssayRouter-C (Table~\ref{tab:support-cv4-strict}). Table~\ref{tab:support-cv4} reports cross-fit accuracy, and the following cost analysis records the subset fits and source calls on the target required to construct it.

The closest reuse methods above are positioning precedents rather than numerical baselines with matched access. Residual compatibility is the profile correlation control within the contract for the explanation suggested by pQSAR. The other original contracts require dense historical prediction profiles, rankings over a model zoo, assay descriptions, training on selected data, or auxiliary experiments, and their actions range from choosing one model to training a new model. Adapting them to frozen heterogeneous outputs, the declared support roles, a fixed four-source action, and the shared NNLS would define a new router. The alternatives within the contract are therefore tested by the coverage matrix in Table~\ref{tab:baseline-coverage}; Support-CV@4 is the primary matched search comparator, while Frozen-DES@4 and MINE-WS@4 deliberately receive richer query access.

\subsubsection{Strict residual compatibility}
\label{app:strict-residual-compatibility}
For each directional partition, a four-fold out-of-fold local target predictor defines $r_{t,i}=\widetilde y_{t,i}-\widehat y_{0,t}^{\mathrm{CF}}(x_i)$ on $R_t$. The profile correlation score is
\begin{equation}
c_t(s)=\operatorname{corr}_{i\in R_t}(\widetilde z_s(x_i),r_{t,i}),
\qquad A_t^{\mathrm{res}}=\operatorname{TopK}_{s\in\mathcal S_t}c_t(s).
\end{equation}
Candidates are ranked by decreasing signed Pearson correlation; the canonical bank order breaks exact ties. The score and ranking use $R_t$ only, while $C_t$ fits the same simplex NNLS with the target predictor and four sources as AssayRouter-C. On the identical 5,184 strict episodes, its NLL is 2.5339, 2.2696, and 2.0806 at 16, 32, and 64 labels, respectively; the overall NLL is 2.2947. The NLL difference between Residual and AssayRouter-C is 0.1844 across the 432 paired target--interface--budget cells, with a two-way target--interface bootstrap interval of $[0.1138,0.2609]$. The corresponding MAE, RMSE, and Spearman values are 32.3550, 44.8218, and 0.3053. This control shares the call budget, access timing, assignments, and final fit while replacing historical post-fit supervision with one correlation rule fitted to the target.

\begin{table}[h]
\centering
\caption{Matched Support-CV@4 cross-fit comparison. NLL is computed after averaging
the same 16 directional prediction vectors. Differences are Support-CV minus
AssayRouter-$\Delta$.}
\label{tab:support-cv4}
\begin{tabular}{lrr}
\toprule
Scope & Support-CV NLL & Difference \\
\midrule
Overall & 2.0448 & 0.0030 \\
16 labels & 2.1806 & 0.0200 \\
32 labels & 2.0315 & 0.0015 \\
64 labels & 1.9223 & $-0.0125$ \\
\bottomrule
\end{tabular}
\end{table}

Support-CV@4 averages 16 $K=4$ directional constituents per episode. It therefore makes 64 source calls, uses 5--21 unique sources per episode (mean 10.45), and has zero episodes with a strict four-source union. The full evaluation performs 20,445,696 inner NNLS fits: 414,720 on Biogen, 8,709,120 on ExpansionRx, and 11,321,856 on TDC. These counts arise during selection to construct the prediction-averaged ensemble; Table~\ref{tab:support-cv4-strict} separately evaluates accuracy under the strict four-call estimand.

The one-shot replay evaluates both methods under the same strict four-call estimand. Support-CV@4 and AssayRouter-C are each scored as 16 independent four-call constituents on the same 5,184 episodes. Table~\ref{tab:support-cv4-strict} gives the paired robust-NLL comparison at every prespecified scope; the complete replay retains the secondary metrics under the same assignments.

\begin{table}[h]
\centering
\caption{Strict four-call Support-CV@4 comparison. Differences are Support-CV NLL
minus AssayRouter-C NLL; positive values favor the frozen historical prior.}
\label{tab:support-cv4-strict}
\begin{tabular}{lr}
\toprule
Scope & Mean difference \\
\midrule
Overall & 0.0409 \\
16 labels & 0.0736 \\
32 labels & 0.0413 \\
64 labels & 0.0078 \\
Biogen & 0.0005 \\
ExpansionRx & 0.0716 \\
TDC & 0.0371 \\
\midrule
ChemBERTa2 / LightGBM & 0.0444 \\
Fine-tuned CheMeleon & 0.0238 \\
Frozen CheMeleon / Ridge & 0.0403 \\
GIN from scratch & 0.0348 \\
Morgan / Ridge & 0.0513 \\
RDKit2D / LightGBM & 0.0505 \\
\bottomrule
\end{tabular}
\end{table}

The right panel of Figure~\ref{fig:central-effect-distributions} shows the target--interface distributions for the matched AssayRouter-$\Delta$ comparison. For AssayRouter-C, the overall difference is 0.0409 with interval $[0.0211,0.0623]$; the 16- and 32-label effects and the ExpansionRx and TDC effects are resolved, whereas the 64-label and Biogen intervals cross zero. AssayRouter-C has lower NLL in 77.8\% of the 144 target--interface units and 63.3\% of the 5,184 support episodes. These rates use the same 16 independently scored four-call constituents as the strict mean.

\subsection{Amortized selection cost}
\label{app:amortized-cost}

\subsubsection{Selection-work proposition}
\label{app:selection-proposition}
Suppose $M$ candidate predictions and their routing features have already been constructed. AssayRouter selects $K$ sources using $M$ evaluations of a frozen scalar score and a top-$K$ operation, with zero counterfactual fits on the target. A sequential exhaustive selector that adds one candidate at a time and fits every remaining candidate requires
\begin{equation}
\sum_{j=0}^{K-1}(M-j)=KM-\frac{K(K-1)}{2}
\end{equation}
counterfactual fits on the target per routing split.

\subsubsection{Proof and boundary}
\label{app:selection-proof}
At step $j$, the sequential selector has already chosen $j$ sources and must evaluate each of the $M-j$ remaining candidates to certify its next greedy choice. Summing over $j=0,\ldots,K-1$ gives the expression above. AssayRouter evaluates its historical score independently for all $M$ candidates and obtains the set by top-$K$. The count isolates exact selection work after frozen predictions and candidate features have been constructed; downstream accuracy, the final NNLS combiner, approximate search, caching, and specialized algebraic updates have separate costs. Support-CV@4 is a stronger screened exhaustive implementation. With $F$ cross-fitting folds, screen width $L$, bank size $M$, and subset size $K$, its selection stage evaluates
\begin{equation}
F\left[M\,\mathbf 1\{M>L\}+{\min(M,L)\choose K}\right]
\end{equation}
counterfactual support fits per direction: the screening term applies when the bank exceeds $L$, and the second exhaustively scores the surviving subsets of size $K$. For the formal comparator, $L=8$ and $K=4$. Both the singleton screen and subset search use only labels and predictions from the routing support, and the selected set is frozen before confirmation scoring; neither post-fit utility nor confirmation labels enter the search. The implementation repeats this operation across targets, interfaces, budgets, episodes, partitions, and role directions. Table~\ref{tab:support-cv4} reports the resulting total of 20,445,696 inner NNLS fits. For TDC, this workload is the same screened procedure over the $L=8$ survivors.

\begin{table}[h]
\centering
\caption{Comparison scopes. Selection fit counts measure subset construction on the target; strict confirmation calls define the accuracy budget. Counts exclude total wall-clock latency.}
\label{tab:selection-cost-ledger}
\begin{tabular}{@{}lrr@{}}
\toprule
Quantity & AssayRouter-C & Support-CV@4 \\
\midrule
Historical label NNLS fits & 28,800 & 0 \\
Selection NNLS fits per direction & 0 & 246.5 \\
Selection NNLS fits, full evaluation & 0 & 20,445,696 \\
Strict confirmation calls per constituent & 4 & 4 \\
Constituents per cross-fit episode & 16 & 16 \\
\bottomrule
\end{tabular}
\end{table}

\subsubsection{Audited fit-count crossover}
\label{app:fit-count-crossover}
Building the complete historical supervision matrix used 711,360 inner NNLS fits; constructing only the standalone labels used by the primary router requires 28,800, followed by 16 HGB fits for model selection and the final model. Support-CV@4 averages 246.5 inner NNLS fits per routing direction, or 3,944 per cross-fit evaluation episode with 16 directions. Measured by this shared NNLS primitive, its repeated search reaches the cost of the primary standalone labels after 116.8 routing directions, equivalent to 7.3 such cross-fit episodes, and the full historical development cost after 2,885.8 directions, or 180.4 episodes. These crossover counts exclude feature construction, HGB execution, frozen source prediction, and the final combiner; they quantify the shared NNLS primitive rather than wall-clock time.

Table~\ref{tab:primary-collection-main} localizes the matched primary comparators by external collection; Table~\ref{tab:primary-interface-main} gives the complementary view across frozen predictor families.

\begin{table}[h]
\centering
\caption{\textbf{Cross-fit greedy and information-rich diagnostics by collection.}
Comparator NLL minus AssayRouter-$\Delta$ NLL, averaged over interfaces and support
budgets; positive values favor the post-fit utility prior.}
\label{tab:primary-collection-main}
\begin{tabular}{lrrr}
\toprule
Comparator & Biogen & ExpansionRx & TDC \\
\midrule
50 random four-source sets & $-0.0026$ & 0.0713 & 0.0275 \\
Support-greedy selection~\citep{caruana2004ensemble} & $-0.0047$ & 0.0547 & 0.0356 \\
Frozen-DES@4 & $-0.0055$ & 0.0013 & $-0.0044$ \\
MINE-WS@4~\citep{moura2021mine} & $-0.0010$ & 0.0421 & 0.0229 \\
All-source convex & $-0.0019$ & 0.0440 & $-0.0274$ \\
\bottomrule
\end{tabular}
\end{table}

\begin{table}[h]
\centering
\caption{\textbf{Cross-fit greedy and information-rich diagnostics by interface.}
Comparator NLL minus AssayRouter-$\Delta$ NLL, averaged over external targets and
budgets; positive values favor the post-fit utility prior.}
\label{tab:primary-interface-main}
\begin{tabular}{lrrrr}
\toprule
Interface & Random & Greedy~\citep{caruana2004ensemble} & DES@4 & All-source \\
\midrule
Morgan / Ridge & 0.0333 & 0.0292 & $-0.0010$ & 0.0078 \\
RDKit2D / LGBM & 0.0445 & 0.0413 & $-0.0005$ & 0.0140 \\
CheMeleon / Ridge & 0.0311 & 0.0304 & $-0.0096$ & $-0.0023$ \\
ChemBERTa2 / LGBM & 0.0474 & 0.0459 & 0.0070 & 0.0100 \\
GIN & 0.0367 & 0.0351 & 0.0026 & 0.0126 \\
CheMeleon / FT & 0.0254 & 0.0141 & $-0.0138$ & $-0.0078$ \\
\bottomrule
\end{tabular}
\end{table}

Table~\ref{tab:hur-permutation-intervals} preserves the marginal-over-states intervention as an independent replication in the same direction as the direct standalone-utility result.

\begin{table}[h]
\centering
\caption{Matched marginal-over-states mechanism tests. Differences are comparator NLL
minus genuine marginal-over-states NLL; positive values favor the genuine labels.}
\label{tab:hur-permutation-intervals}
\begin{tabular}{lr}
\toprule
Comparator and scope & Mean difference \\
\midrule
State-conditioned extension, overall & 0.0011 \\
State-conditioned, 16 labels & 0.0051 \\
State-conditioned, 32 labels & $-0.0012$ \\
State-conditioned, 64 labels & $-0.0004$ \\
\midrule
Marginal labels permuted, overall & 0.0532 \\
Marginal labels permuted, 16 labels & 0.0824 \\
Marginal labels permuted, 32 labels & 0.0428 \\
Marginal labels permuted, 64 labels & 0.0344 \\
Marginal labels permuted, Biogen & 0.0006 \\
Marginal labels permuted, ExpansionRx & 0.0716 \\
Marginal labels permuted, TDC & 0.0698 \\
\bottomrule
\end{tabular}
\end{table}

\section{External Checkpoint and Benchmark Reproductions}
\label{app:external-reproductions}

\subsection{Checkpoint and benchmark reproductions}
\label{app:benchmark-reproductions}

\subsubsection{Official few-shot checkpoints}
\label{app:few-shot-checkpoints}
We downloaded the no-FEP ChEMBL checkpoints released with ActFound~\citep{feng2024actfound} and evaluated five official systems (ActFound, ActFound-transfer, MAML, ProtoNet, and transfer-QSAR) on the nine TDC regression targets. Every method receives the identical 16/32/64 support molecules used by the corresponding AssayRouter episode and predicts the same official test split. The official 2,048-dimensional count-Morgan input and five-step adaptation are preserved. Across means computed over targets, the strongest official checkpoint is ProtoNet: the NLL difference between AssayRouter and ProtoNet is $-0.0190$, $0.2182$, and $0.0125$ at the three budgets, where positive values favor ProtoNet. Bootstrap intervals over targets cross zero in all three cases; the complete matrix for all five methods accompanies the submission.

\subsubsection{Published Biogen predictors}
\label{app:biogen-reproductions}
We also reproduce methods evaluated on the exact public Biogen~\citep{fang2023prospective} split. The original Random Forest~\citep{breiman2001random} and LightGBM~\citep{ke2017lightgbm} code gives mean Pearson correlations of 0.6683 and 0.7069, close to the published 0.6700 and 0.7000. The official MolSetRep~\citep{boulougouri2024molsetrep} GINE implementation completes three runs for each of six endpoints and reaches mean Pearson $r=0.6373$; its SR-GINE counterpart reaches $r=0.7259$ in another 18 runs. Across both neural models, endpoint values reproduce the published table within 0.029. These are full-training predictors and are reported separately from support-budget-matched routing methods. Table~\ref{tab:biogen-full-training} gives the resulting macro correlations and run counts.

\begin{table}[h]
\centering
\caption{Official full-training reproduction on all six public Biogen endpoints.
Pearson $r$ is macro-averaged across endpoints.}
\label{tab:biogen-full-training}
\begin{tabular}{lrr}
\toprule
Method & Runs & Macro Pearson $r$ \\
\midrule
Random Forest~\citep{breiman2001random} & 6 & 0.6683 \\
LightGBM~\citep{ke2017lightgbm} & 6 & 0.7069 \\
MolSetRep GINE~\citep{boulougouri2024molsetrep} & 18 & 0.6373 \\
MolSetRep SR-GINE~\citep{boulougouri2024molsetrep} & 18 & \textbf{0.7259} \\
\bottomrule
\end{tabular}
\end{table}

\subsubsection{MiniMol reproduction}
\label{app:minimol-reproduction}
The official MiniMol~\citep{klaser2024minimol} protocol uses all 22 TDC ADMET tasks, train/validation splits from five author-provided seeds, five fold heads per split, and 25 epochs per head. This yields 550 fitted heads above one frozen official checkpoint. All 22 tasks complete; the mean and maximum absolute deviations from the released metrics are 0.0122 and 0.0830, respectively. The complete task-level comparison accompanies the submission.

\subsubsection{Contrastive KERMT reproduction}
\label{app:kermt-reproduction}
We fine-tune the official Contrastive KERMT~\citep{xue2026contrastivekermt} checkpoint for 100 epochs on each of five seeds and retain the authors' multitask readout for each task. The reproduction completes four Biogen endpoints with high coverage and all nine ExpansionRx endpoints, giving mean absolute error (MAE) averaged over endpoints of 0.3515 and 0.2836, respectively. Table~\ref{tab:kermt} separates these public-checkpoint reproductions from the manuscript's reported macro means; both use full training and remain separate from routing at matched support budgets.

\begin{table}[h]
\centering
\caption{Full-training KERMT comparison (macro MAE; lower is better). Author values
are transcribed from the official manuscript source; reproductions use the released
Contrastive KERMT checkpoint, five seeds, and the frozen public splits.}
\label{tab:kermt}
\begin{tabular}{lrr}
\toprule
Protocol & Biogen (4) & ExpansionRx (9) \\
\midrule
Author KERMT task-specific & 0.3320 & 0.3750 \\
Author best Contrastive KERMT & \textbf{0.3210} & 0.3590 \\
Official-checkpoint reproduction & 0.3515 & \textbf{0.2836} \\
\bottomrule
\end{tabular}
\end{table}

Target-only and all-source fit the same regularized combiner on the local prediction or the full bank. Compatibility, training-size, and signal-to-uncertainty selectors rank the frozen bank by their named support statistic. Local-kernel estimates residual compatibility among molecular neighbors. QComp-style~\citep{yang2025qcomp} applies a residual-covariance update through the same interface; original QComp can observe auxiliary experimental values for a query and therefore has a different information budget.

\subsubsection{Statistical and reproducibility details}
\label{app:full-training-statistics}
\label{app:statistics}

Let $r=(y-\hat y)/\sigma_{\mathrm{sup}}$, where $\sigma_{\mathrm{sup}}$ is the median absolute deviation of the complete episode support $R\cup C$, multiplied by 1.4826. The sample standard deviation is used when that quantity vanishes, with a $10^{-6}$ floor. Robust NLL is the Student-$t$ loss for point predictions with $\nu=3$,
\begin{equation}
\ell_{\mathrm{NLL}}(r)=\tfrac12\log(\nu\pi)
+\log\Gamma(\nu/2)-\log\Gamma((\nu+1)/2)
+\tfrac{\nu+1}{2}\log(1+r^2/\nu).
\end{equation}
The methods do not emit predictive variances for this metric. We average support repetitions before comparison. Aggregate bootstraps independently resample predictor interfaces and target tasks, preserving their crossed structure. Each analysis uses at least 4,000 replicates; intervals are reported with the corresponding results.

All controls reuse identical supports, predictions, combiners, and evaluation rows. Random subsets are fixed before scoring, and confirmation labels never participate in selection. The supplementary release preserves aggregate metrics and the method contracts needed to audit the reported comparisons.

\section{Retrospective Development Boundary}
\label{app:retrospective-development}

Earlier sourcewise studies used ExpansionRx, Biogen, and TDC to develop the abstraction of prediction contracts and the fixed four-source action. They established three design choices carried into the reported protocol: residual behavior is sufficient to expose portable structure between source and target; a common convex combiner is necessary to compare selectors without confounding from coefficient fitting; and a fan-in of four channels balances complementary corrections against interference from the full bank. Their scientific role is selection of the deployment contract.

AssayRouter in the main comparisons is the HGB regressor without identity features defined by Equations~\ref{eq:set-utility}--\ref{eq:router-objective}, trained on standalone post-fit utility centered within each block and frozen before the action in Equation~\ref{eq:one-shot-routing}. Marginal-over-states and state-conditioned regressors are matched refinements. Earlier pairwise, soft-prior, graph, classification, and semantic systems remain separate development studies.

The external claim is model-held-out transfer under this fixed contract: all 24 evaluation targets are excluded from router fitting and grouped capacity selection. Appendix~\ref{app:set-conditional-protocol} reports the evidence through frozen candidate-label interventions, leave-one-interface-out routers, strict matched comparisons, identity holdouts, provenance checks, and leave-one-cluster sensitivities. These tests carry the paper's conclusions; the retrospective studies explain how the final question and action budget were chosen.

\end{document}